# Octopuses: biological facts and technical solutions

Oxana Shamilyan[1], Ievgen Kabin[1], Zoya Dyka[1], Michael Kuba[3] and Peter Langendoerfer[1,2]
[1] IHP – Leibniz-Institut für innovative Mikroelektronik, Frankfurt (Oder), Germany
[2] BTU Cottbus-Senftenberg, Cottbus, Germany
[3] Okinawa Institute of Science and Technology Graduate University (OIST), 1919-1 Tancha, Onna-son, Okinawa 904-0945, Japan
{shamilyan, kabin, dyka, langendoerfer}@ihp-microelectronics.com

*Abstract*—**Octopus is an invertebrate belonging to the class of *Cephalopoda*. The body of an Octopus lacks any morphological joints and rigid parts. Their arms, skin and the complex nervous system are investigated by a several researchers all over the world. Octopuses are the object of inspiration for my scientists in different areas, including AI. Soft- and hardware are developed based on octopus features. Soft-robotics octopus-inspired arms are the most common type of developments. There are a lot of different variants of this solution, each of them is different from the other. In this paper, we describe the most remarkable octopus features, show solutions inspired by octopus and provide new ideas for further work and investigations in combination of AI and bio-inspired soft-robotics areas.**

## I. INTRODUCTION

The most recent vision concerning cyber-physical systems (CPS) or cyber-physical systems of systems (CPSoS) is to empower them to become resilient. This requires them to be at least:

- self-aware,
- robust to fluctuation of environmental parameters (or self-adaptable),
- resistant against malicious manipulation,
- intelligent to find themselves a correct solution in complex and non-expected situations.

All these capabilities require that the system is on the one hand empowered by sensors to learn about its own state and its environment and on the other hand that it is equipped with means to interpret the data collected and to determine how to adapt its behaviour or event itself to changes detected. Most biological "systems" at least those assumed to be on a higher level of development have these capabilities, which is the reason why often engineers rely on inspiration of biological systems. The term "bio-inspired" means that solution is taken from biosphere and mimics nature. The architecture normally is based on animals´ behaviour, their lifestyle, ways of making decision, interaction with surroundings, etc. Therefore, researchers achieve a lot of different results, because expansion of the study area allows to looking at the problem from other perspective.

In our case we are aiming at learning and adapting resilient systems based on the fascinating capabilities of Octopus. They are smart, show unusual behaviour and have remarkable features, like body arrangement, embodied intelligence and distributed nervous system. Many researchers and researcher groups are inspired by them. There are many scientific studies, inspired by the octopuses, in different areas: biological researches, robotics, soft- and hardware developments, investigation and implementation of AI methods. The octopus projects encompassing groups of researchers are very common as well.

The scope of materials is huge, but more than the half of them are biological papers. Robotics solutions are the ones that are mostly researched in the field of engineering. But there are also software solutions and system architectures inspired by Octopus. The least part is the investigation in the field of AI despite this was the first field that used Octopus as inspiration or at least to verify own ideas.

The aim of this paper is to show known octopus benefits and describe ideas of further investigations in the intersection of octopuses and artificial intelligence and resilience.

This work is based on 70 science articles we found performing an exhaustive search of papers published in the last the 20 years. Table I provides numbers of papers evaluated in this paper sorted by topics.

TABLE I. TOPICS DISCUSSED IN REFERENCED HERE ARTICLES

| Article topic | Bio | Robotics | Software | AI | Projects | Total |
|---|---|---|---|---|---|---|
| Quantity | 38 | 17 | 5 | 5 | 5 | **70** |

In order to ensure self-containment of this paper we will introduce biological facts of Octopus in some detail. On the one hand this aims at providing common ground but on the other hand we hope that these fascinating feature will trigger more ideas how to improve technical systems.

The rest of this paper is structured as follows. First, as mentioned already we introduce the Octopus as the source of inspiration. Then we introduce technical systems fully or partially inspired by Octopuses.

## II. OCTOPUSES: BIOLOGICAL FACTS

Octopuses now are one of the most popular cephalopods thanks to their interesting morphology, features of their soft body structure with a bilaterally symmetric form with eight arms and their evolved cognition. Fig. 1 shows the octopus morphology. Octopuses belong to the class of Cephalopoda. They are invertebrates with lack of any rigid part [1]. Octopuses live in every ocean and most species spend a significant amount of their time in their dens.

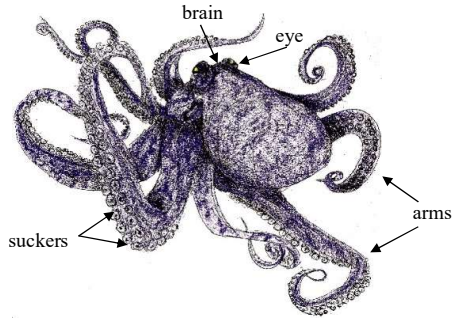

Figure 1. Octopus morphology

In average octopus lives about 12-18 months. Females usually die soon after hatching their eggs [2], [3]. Young individuals do not have parents care and always scattering right after hatching. Many octopuses spend their life solitary and limit contacts with other octopuses. Many species are described as non-social and solitary animals. This means that the individuals are gathering their fascinating skill almost on themselves which indicates even more cognition as if they would be taught from older conspecifics and thus learn from adult octopuses.

Octopuses are predators and feed on other invertebrates like clams, snails, crabs and prawns they also hunt vertebrates like small fish. As they are predators, but not on the top of the food chain, in different marine inhabitants octopuses are prey as well. Cannibalism is also found among cephalopods [2], [4]. They are soft-bodied and can become an easy prey for predators, likely because of this they encompass a lot of features to avoid it. They include body mimicry and anti-predatory strategies. For example, octopuses can shoot black ink with a water jet creating a dark non-transparent, olfactory disturbing cloud allowing them to escape [5]. Octopuses are smart enough to use different tools as protection, for example cocoanut shells and stones [5], [6].

Octopuses, squids and cuttlefishes have three hearts: two branchial hearts and one systemic heart, in addition to these 3 hearts several vessels of the closed blood system are muscular and help to move the blood trough the soft body. The branchial hearts help oxygenate blood in the gills and are connected to the systemic heart, which pumps blood to the whole body [7].

The one of the most remarkable features of octopus is nervous system. Overall it counts 300-500 million nerve cells for adult common octopuses. Their complex nervous system and their cognition make them one of the most intelligent of all vertebrates [1], [2]. For comparison a dog has about 600 million nerve cells. The neural system of octopus has a unique structure. The most interesting feature is that the octopus intelligence is not centralized, and have highest brain-to-body mass ratio within all invertebrate. The whole nervous system is divided into two groups: a central nervous system (CNS) and the peripheral nervous (PNS) system [1], [5], [8].

The neural system is the most important component of the sophisticated octopus behaviour and cognition. Many research works are studying how octopuses interact with the environment, make a decisions, etc. The following four main components can be distinguished here: sensory perception, perception-based cognition, learning, and memory. Each of these groups has a significant impact on the complex octopus cognition [9].

Octopus vulgaris has fewer neurons in the CNS than in its arms. The octopus CNS supports an acute and sensitive vision system, good spatial memory, decision-making, and camouflage behaviour [3]. Two large optic lobes are connected to the retinae of the camera-like eyes [1]. The PNS in the arms accounts for almost two third of all neurons that means each arm possesses about 40 million neuron cells. Each arm has an autonomous individual structure so it can act alone or in coordination with other arms without coordination or control from the CNS. To understand the inner structure of the brain several studies were conducted by researchers for the last 150 years. More recently scientists started to use neurophysiological methods and modern imaging techniques to better understand the neurological procedures and the connection of the different brain sections [10], [11], [12], [13].

The study of the octopus cognition structure shows that all components of the octopus body are connected with each other. Neuron signals are transmitted from brain to arms to implement some actions, from arms to brain to process sensory information, from eyes to skin to change appearance and so on [14]. The whole body works as a single mechanism and implements various complex tasks.

Octopuses are intelligent, soft-bodied animals with keen senses that perform reliably in a variety of visual and tactile learning tasks. They are capable of solving problems such as: opening a screw top jar, using coconut halves as protection [3], moving an object rough a hole in a wall and finding prey in a transparent maze [15].They are able to learn and have a short- and a long- term memory. The evidence of octopus memory was presented in [16]. The research proves that cephalopods possess "What-Where-When" memory – 'what' (prey type) was located, 'where' (location of the visual cue) and 'when' (time elapsed since they previously ate). In the wild, octopuses tend to avoid places where they have fed in previous days [2], so they use their brains and memory to maintain the productive hunting "win switch" strategy.

A thorough understanding of how octopuses receive information from their environment and process it can be very helpful knowledge that can be replicated in implementing of AI methods.

Researches show that Octopus vulgaris can use vision to learn, i.e. they have the ability for observation learning [3], [15], [17]. Octopus can learn to choose correct objects, how to find the place with food, which place of feeding is correct and which is not. Octopuses are rapidly learning by observing. For example, training Octopus vulgaris to choose a correct ball needed about 4 trials. Right after the training phase octopuses choose the correct ball in 85% of all trails. The result remains approximately the same (81%) even 5 days after training [16], [18].

As mentioned before, octopus is a soft-bodied animal without any rigid external or internal skeleton. They use their boneless to change their shape and taking different forms. This ability allows them to squeeze into different narrow holes [5].

The lack of a skeleton is compensated by a muscular structure called "muscular hydrostat". Muscles in the octopus body can easily stiffen, relax, elongate or shorten. This allows octopuses to solve different tasks, explore items and use tools by their arms. Each arm is independent and can be monitored by itself. If an arm is damaged, it can be regenerated. The size and force of regenerated arm, depends on the age of the octopus [19]. Amputated arm is still active within an hour after amputation, it continues to bend, makes same movements as usual, grabs the objects by the suckers [1], [4]. This behaviour confirms distributed structure of nervous system. Arms do not need to be attached to the central brain to make stereotyped movements. The big number of nervous cells might provide a memory ability and allows to arms "be alive" for a next hour even without receiving the nerve impulses from the brain.

Octopus arms represent a high flexible and multifunctional instruments.. Octopus arm encompasses three types of muscles. Each of the type is responsible for a particular action. For example, extending of the longitudinal muscles leads to the arm-elongation and stiffness of the whole arm depends on the transverse muscles. The nerve cord runs along whole arm and has a sinusoidal arrangement that allows elongating and bending of an arm without stretching the cord itself [20]. These arms have potentially infinite degrees of freedom, can bend in any direction and twist either clockwise or counter clockwise. The general arm movements are complicated. However, for grabbing objects octopuses use rather simple and predictable locomotion strategy and stereotyping movements [1]. Octopus uses their arm either for locomotion – swimming, walking on the sea bottom, crawling – for reaching, grasping, digging and grooming. While crawling arms push the body by elongating thus the direction of the arms is opposite to the direction of body movement [1]. These methods of locomotion were taken as an example in development of artificial octopus [21]. Researches show that octopuses prefer to use particular arms (or pairs of arms) while crawling [22]. Usually octopuses prefer crawling in 45 degrees [1], [22]. This direction of crawling is quite usual for animals with bilateral symmetry in the body. In addition, it provides more efficient work of their sensory system that means it helps perceive the environment better. Octopuses use their arms for changing the direction of swimming. Using different combinations of arms or pairs of arms allows them to change the angle and speed of movements. Arms are also used to seize, manipulate, and bring food to the mouth [4].

Another interesting part of octopus arm are the suckers. The octopus' pelvic side of the arms can have up to hundreds of suckers. Each sucker contains a large number of tactile and chemical receptors [4]. The suckers are a sophisticated organ. It can attach to objects with varying degrees of force. This includes anchoring to perfectly smooth surfaces as well as to surfaces with a certain roughness [23]. Suckers apply significantly different forces to different types of objects and surfaces, but the suckers neither grasped their own arm, because of a skin self-recognition mechanism [4], [24], even if the arm was amputated.

Cephalopods can change the skin colour and texture very quickly i.e. within 270–730 milliseconds even though being colour blind, but sensitive to polarised light [2], [25]. Their skin is a perfect tool for anti-predatory strategy as it provides an ideal means for camouflage, to make cephalopods difficult to detect in their surroundings, or, in opposite, become very bright, to make predators think they are venomous.

The skin is divided into the five layers of different optical elements – three layers of chromatophores, mirror-like iridophores, and matte white leucophores [26].

The chromatophore system encompass millions of pigment cells called chromatophores. Each chromatophore cell contains an elastic sac with the pigment, which is opened by muscles attached to the cell. Changing the colour depends on situation and surroundings [27]. This process can be represented as a sequence of neural signals (see Fig. 2) [28].

The colour changing is directly connected with information received from the eyes and appears as a product of polarization vision [29]. If chromatophore sacs are narrowed, then the light reflects in iridophores which give a good camouflage at low light levels [19].

Octopuses are able to mimicry. A good example of this ability is described in [30]. The mimic octopus (*Thaumoctopus mimicus*) can imitate behaviour and skin pattern of a sea-snake lion fish or other toxic and potentially dangerous animals. This colour changing is used as an anti-predatory strategy.

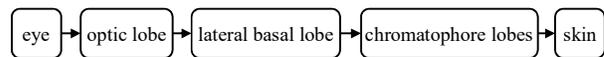

Figure 2.  Sequence of neural signals from eyes to skin

Octopuses use a lot of different patterns in their skin painting. The most common is so called «passing clouds» that looks like a travelling waves on cephalopods skin [31]. Skin colour-patterns also depend on the cephalopod species, for example *Larger Pacific Striped* Octopus is striped and uses his skin to mimic the see snake as *Thaumoctopus mimicus* Octopus does [32].

Cephalopods eyes have one unusual property called visual lateralization. Visual lateralization allows to prioritize information from one visual field over the other one. In addition one side of the cuttlefish brain is specialized on processing information used for hunting, and the other side is used for monitoring area to occurring predators [25], [33]. Eyes are also used for body- and arms-positioning and to locate food [17].

## III. OCTOPUS-INSPIRED SOLUTIONS

External and internal parts of the octopus body are also remarkable not only from biological perspective, but also for technical development. A lot of works and researches are made within projects [34], [35], [36], [37], [38] and within a PhD thesis [39], [40]. The next subsection give insight into how octopus features can be used in technical developments.

### A. Nervous system

The nervous system represent an unusual organization. The central brain acts like a decision-making mechanism, it initializes and pulls the impulse to the peripheral. However brain does not issue top-down commands for every small movement of the arms [41], a lot of decisions arms make by themselves.

The scientists in The Octopus project [42] are focused in study of octopus brain functions. Their aim is to understand the behaviour of octopuses and how the octopus brain works. Using this information, they want to create a model to simulate a distributed intelligence of octopuses.

A similar research is done at University of Nevada, Reno. They use an electroencephalography sensor to record nervous signal. After analysis they plan to divide all this record to its simplest parts. These primitives will be helpful in understanding of how octopus brains work and in development of octopus-like nervous system as well [38].

The structure of octopus nervous system and the way how brain communicates with arms can be used in software architecture developments. For example, [43] demonstrates the developing process and application of multiprocessing framework with a core based on octopus nervous system arrangement. The main idea is to distribute control between main processor and coprocessors, to make coprocessors solve small tasks by their own and to increase the processing speed of whole system.

An additional example of an Octopus inspired software development is described in [44]. The group of scientists provides a model of a system that detects the malicious node and drains its battery using the other nodes of network i.e. the nodes start to send the messages (flooding) to malicious node, just as the octopus spread the ink to predators.

Artificial intelligence (AI) is an interesting field of research providing powerful and flexible mechanisms suited for different application fields such as – analytics, agriculture, medicine, robotics, software development. Despite the huge advancements of AI in recent years due to increased computing power there is still a huge perspective for further improvements and developments. In the field of artificial intelligence, the exchange of ideas between scientists in the technical and biological fields is extremely fruitful [45]. The comparatively simple 2-layer neural network of octopuses corresponds to the first artificial networks that could learn classifications. AI methods are often used in robotics. Different AI algorithms can be used in robotics. [46] provides an overview of major AI paradigms such as the hierarchical paradigm and the reactive paradigm. Furthermore, it includes examples how to transfer principles of the paradigm into a real robotic solution. Robots become more intelligent, adjust behaviour algorithms, learn how to perform the tasks, and then are able to apply this knowledge in real situations. Data are collected from sensors and processed by machine learning algorithms [47]. Usually AI methods were inspired from human intelligence. Recently, the focus is shifted from human to animal intelligence.

### B. Arrangement of body and arms

There are two types of bio-inspired robots namely soft-robots and rigid ones. The later are usually built by rigid modules linked with each other by rigid joints and are wide used in industry. Such robots increase efficiency of factory work and improve is quality. They perform the tasks with higher accuracy and don't need breaks as humans do. However, they also have a major disadvantage. Rigid robots are placed in known environment or even sometimes the environment is built regarding the robots dimensions and other specifics. This type of robots loses its advantages to a certain extend or even became useless in unexplored and unprepared areas. Here soft-robots are by far more promising. These bio- solutions imitate a whole body of animal or some parts of their bodies. There are developments of insect, aquatic and amphibious animals and primates. In more particular, there are soft-robotics inspired by worms, fish, elephant trunk, octopus arm, caterpillars, etc. [48], [49]. The main advantages of soft-robots is their lack of rigid modules. They are usually made of soft materials such as gels, rubber or silicone. This soft structure allows them to squeeze into hard-accessible areas. Also, the soft-robots can grasp and manipulate the objects without them.

Octopus arms are source of inspirations for many researchers. Their main purpose is to build robots from a soft materials – soft-robots. Such robots have a list of advantages compared to rigid robots. First of all they are highly resistant to stress forces, as they don´t have rigid links, can be used in different areas with unexplored environment and can bend producing a smooth curves [50]. The soft structure of these robots allows to use them as search or rescue robots, which could crawl into a hard-accessible area and explore it [51].

A large part of soft-robotics solutions and octopus investigations were made by the group of researchers from BioRobotics Institute, Scuola Superiore Sant'Anna, Pisa, Italy within The OCTOPUS project [35]. For example, [20] includes description of instruments developed and measuring protocols to obtain measurement of octopus arm inner structure. The group of researchers made mechanical measurements of significant properties of octopus arm such as: elongation, strength, shortening and longitudinal stiffness. They also did morphological measurement of transverse and longitudinal muscles, sinusoidal nerve cord and density of the arm. The biomechanical and anatomical investigations were used in prototype development in the second part of research [52]. The development process and final solution of silicone arm are shown, which are based on the measurements and the morphological features obtained in their previous work. In [53] their prototype was improved and tested in the water area. Octopuses control their arms thanks to mechanism of embodied intelligence, what´s make them a good prototype for including artificial intelligence in soft robotic arms [53]. The example of octopus control system representation shown in Fig. 3.

A soft arm prototype with locomotion and grasping abilities was developed by the same group. The prototype has a conic shape made of silicone with one steel and one fibre cable inside. The steel cable allows the arm to elongate and shorten providing locomotion. The fibre cable allows the arm to bend and grasp the objects. Both cables are actuated by servomotors [50]. At the end an eight-arm octopus-like robot was constructed. It is fully functioning under the water, using a remote control, , due to its soft arm materials and conical shape it can grasp objects of different size and shape [54].

Another soft-robotic octopus arm was developed by Researchers at the Harvard John A. Paulson School of Engineering and Applied Sciences (SEAS) and Beihang University. The soft-robotic arm encompasses all main features of real octopus arm and mimicked the general structure. It can grip, move, and manipulate a range of different objects with its flexible and tapered design [55].

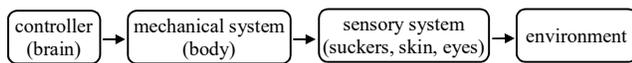

Figure 3. Octopus control system

The development of soft robot arm-prototypes requires as much similarity with real octopus arms as possible. This is the reason why researches need to test different texture and materials. In [56] a platform for testing mock-ups was designed and developed. It helps to reveal all advantages and disadvantages of using the arm prototype.

The development of an octopus-like arm it is just a first step. The main idea of a lot of scientific works is to empower such arms to move and act like a real octopus arm. To achieve this real octopus movements were analysed. [57] describes the process of recording octopus movements, dividing them into primitives, clustering and classifying them. The information provided can be used as in further investigations e.g. when developing octopus-like locomotion strategies for artificial prototypes. Octopus motor control is a sophisticated system, but it can be investigated, divided into primitive parts and used in octopus-inspired prototypes [11].

Also a group of researchers from Michigan state university analysed the movements of octopus arms using video-recording and artificial intelligence. Data was collected by electrodes implanted in the octopus' arms. The analysed information can be helpful in development of intelligent sensors [34].

Machine learning algorithms and especially the reinforce learning are commonly used for training octopus-like manipulators. In particular, the reinforce learning is used to train octopus-like manipulator arm to relocate objects from the start to the goal position [58].

The complexity of the learning depends only on the number of degrees of freedom, and not on the size of the robotics [59].

There are several technical solutions that are inspired by octopus suckers even though they did not gain as much popularity as those inspired by the complete octopus arm. The idea is develop artificial suckers that can attach to different surfaces. The first development of artificial suckers was presented in [60]. [23] reports about improvements especially and a sensory unit was added, for achieving better sensing of the proximity and tactile interactions.

Besides the octopus-like suction cups octopus-inspired sensor sticks [61] were developed. The group of researchers developed graphene-based adhesive bio-sensor that looks like artificial skin. It sticks to dry or wet human skin. The inspiration was taken of octopus suckers and how they work in air and water environment. However, soft robotics arm are usually developed without suckers, probably in order not to overload the development.

*C. Skin, mimicry*

Some works are focused on skin genome researching [62], [63]. These researches help to understand the internal process of appearance changing. These works discovered that eyes and skin have the same structure [64] and can detect light intensity, and change the colour according to light changings [65]. The results of the genome studies enables the creation of artificial skin and skin-cells with all appropriate features [66], [67], [68].

The main topic of investigation and development is to mimic the ability of cephalopods skin. The scientist are focused on developing an artificial skin with the ability to change colour regarding to surroundings. Based on the results of octopus skin and researching its genome the artificial skin was developed [66]. It can mimic the touch effect and provides similar behaviour and patterns as octopus skin does.

Another example of skin developments is described in [69]. There a high-performance system consisting of arrays of actuators and photodetectors is reported. The developed skin has a multilayer architecture with embedded artificial chromatophores. Its flexible design allows to attach the unit itself to some object to provide camouflage.

A 3D printed octopus-inspired camouflage robot was developed by researchers from Rutgers University [70]. The robot is covered with ´smart gel´ including artificial chromatophores. The gel composition turns light into heat energy and controls the size of artificial chromatophores sacs.

IV. IDEA FOR FURTHER WORKS

According to the above overview, it becomes clear that the most of the researches were done in the field of biology. The second largest group of papers deals with investigation in the field of robotics focusing on octopus-like arm development. Soft-robots are candidates for performing complex tasks in an unknown environment if they are resilient. Only a small part of all reviewed works are somehow related to resilience or AI. That means this part is the less explored and can become a good area for new investigations.

AI is the main mean to achieve self-adaptability which is kind of the ground truth when it comes to achieve resilience of a system. The idea for further works and researches is an application of artificial intelligence methods to octopus-inspired soft-robots locomotion. Soft-robots usually are used in unknown area. Hence, they need to be resilient enough to adapt in the unknown environment. So the aim is to make a soft-robotics arm to move by itself without remote control. The decisions how and in which direction to move shall be taken based on the data obtained from the sensors exploring the surroundings. Information provided in biological papers points out that octopus movements can be predicted. That´s why the idea of autonomous movements generated by AI methods and machine learning techniques seems rather possible.

Octopus-inspired soft-robotics arms can be used in different fields. They can be helpful in the surgeries or in the rescue operations. They can squeeze in narrow holes that make them a useful tool. The sensors can check various parameters and provide information about the surrounding area. Based on this information autonomous movements of soft-robotics arm can be generated.

## V. Conclusion

Octopuses are one of the most interesting species. They are interesting not only for biological researches, but also for technical development. Their external and internal structures are unique and can be used as models in different developments, first of all in robotics. Artificial intelligence is a powerful tool, suitable for a lot of cases. Using artificial intelligence in robotics is already common practice, but both of these areas are continuously improving. Our idea aims at combining these areas. Octopus like soft-robots are already a wide spread concept of development. AI methods are rapidly developing, changing and improving, what makes them flexible and useful. They are used in very different areas, improving the speed and quality of work. Our idea is to use AI methods in soft-robotics to increase their independence of human decisions and their resilience in unexplored environments.

References

[1] G. Levy, N. Nesher, L. Zullo, and B. Hochner, "Motor Control in Soft-Bodied Animals,", The Oxford handbook of invertebrate neurobiology, J. H. Byrne, Ed., New York: Oxford University Press, 2019, pp. 494–510.

[2] S. S. Adams and S. Burbeck, "Beyond the Octopus: From General Intelligence Toward a Human-Like Mind," in Atlantis thinking machines, Kai-Uwe Kuhnberger, vol. 4, Theoretical foundations of artificial general intelligence, Paris: Atlantis Press, 2012, pp. 49–65.

[3] P. Amodio, et al. "Grow Smart and Die Young: Why Did Cephalopods Evolve Intelligence?," Trends in ecology & evolution, vol. 34, no. 1, pp. 45–56, 2019.

[4] N. Nesher, G. Levy, F. W. Grasso, and B. Hochner, "Self-recognition mechanism between skin and suckers prevents octopus arms from interfering with each other," Current biology, vol. 24, no. 11, pp. 1271–1275, 2014.

[5] M. Jukic, "It's Time to Take Octopus Civilization Seriously," Palladium, 02-Apr-19. https://palladiummag.com/2019/04/01/its-time-to-take-octopus-civilization-seriously/ (accessed: 12-Mar-21).

[6] A. K. Schnell and N. S. Clayton, "Cephalopod cognition," Current biology: CB, vol. 29, no. 15, R726-R732, 2019.

[7] E. Maldonado, E. Rangel-Huerta, R. González-Gómez, G. Fajardo-Alvarado, and P. S. Morillo-Velarde, "Octopus insularis as a new marine model for evolutionary developmental biology," Biology open, vol. 8, no. 11, 2019.

[8] "Do octopuses' arms have a mind of their own? Researchers are unravelling the mystery of how octopuses move their arms,". ScienceDaily. [Online]. Available: https://www.sciencedaily.com/releases/2020/11/201102120027.htm (accessed: 12-Mar-21).

[9] A. K. Schnell, P. Amodio, M. Boeckle, and N. S. Clayton, "How intelligent is a cephalopod? Lessons from comparative cognition," Biological reviews of the Cambridge Philosophical Society, vol. 96, no. 1, pp. 162–178, 2021.

[10] W.-S. Chung, N. D. Kurniawan, and N. J. Marshall, "Toward an MRI-Based Mesoscale Connectome of the Squid Brain," iScience, vol. 23, no. 1, p. 100816, 2020.

[11] T. Gutnick, T. Shomrat, J. A. Mather, and M. J. Kuba, "The Cephalopod Brain: Motion Control, Learning, and Cognition," in Physiology of molluscs; a collection of selected reviews, vol. 2, Apple Academic Press, 2016, pp. 137–177.

[12] N. Nesher, F. Maiole, T. Shomrat, B. Hochner, and L. Zullo, "From synaptic input to muscle contraction: arm muscle cells of Octopus vulgaris show unique neuromuscular junction and excitation-contraction coupling properties," Proceedings. Biological sciences, vol. 286, no. 1909, p. 20191278, 2019.

[13] L. Zullo, H. Eichenstein, F. Maiole, and B. Hochner, "Motor control pathways in the nervous system of Octopus vulgaris arm," Journal of comparative physiology. A, Neuroethology, sensory, neural, and behavioral physiology, vol. 205, no. 2, pp. 271–279, 2019.

[14] J. A. Mather and L. Dickel, "Cephalopod complex cognition," Current Opinion in Behavioral Sciences, vol. 16, pp. 131–137, 2017, .

[15] T. Gutnick, R. A. Byrne, B. Hochner, and M. Kuba, "Octopus vulgaris uses visual information to determine the location of its arm," Current biology: CB, vol. 21, no. 6, pp. 460–462, 2011.

[16] C. Jozet-Alves, M. Bertin, and N. S. Clayton, "Evidence of episodic-like memory in cuttlefish," Current biology, vol. 23, no. 23, R1033-5, 2013.

[17] J. E. Niven, "Invertebrate neurobiology: visual direction of arm movements in an octopus," Current biology, vol. 21, no. 6, R217-8, 2011.

[18] G. Fiorito and P. Scotto, "Observational Learning in Octopus vulgaris," Science (New York), vol. 256, no. 5056, pp. 545–547, 1992.

[19] J. Mather, "What is in an octopus's mind?," Animal Sentience, vol. 4, no. 26, 2019.

[20] L. Margheri, C. Laschi, and B. Mazzolai, "Soft robotic arm inspired by the octopus: I. From biological functions to artificial requirements," Bioinspiration & biomimetics, vol. 7, no. 2, p. 25004, 2012.

[21] Y. Sakuhara, H. Shimizu, and K. Ito, "Climbing Soft Robot Inspired by Octopus," in 2020 IEEE 10th International Conference on Intelligent Systems (IS): Proceedings, Varna, Bulgaria, 2020, pp. 463–468.

[22] G. Levy, T. Flash, and B. Hochner, "Arm coordination in octopus crawling involves unique motor control strategies," Current biology, vol. 25, no. 9, pp. 1195–1200, 2015.

[23] S. Sareh et al., "Anchoring like octopus: biologically inspired soft artificial sucker," Journal of the Royal Society, Interface, vol. 14, no. 135, 2017.

[24] R. J. Crook and E. T. Walters, "Neuroethology: self-recognition helps octopuses avoid entanglement," Current biology, vol. 24, no. 11, R520-1, 2014

[25] A. K. Schnell, C. Bellanger, G. Vallortigara, and C. Jozet-Alves, "Visual asymmetries in cuttlefish during brightness matching for camouflage," Current biology, vol. 28, no. 17, R925-R926, 2018.

[26] T. L. Williams, et al., "Dynamic pigmentary and structural coloration within cephalopod chromatophore organs," Nature communications, vol. 10, no. 1, p. 1004, 2019.

[27] D. Osorio, "Cephalopod behaviour: Skin flicks," Current biology, vol. 24, no. 15, R684-5, 2014.

[28] R. Hanlon, J. B. Messenger "Adaptive coloration in young cuttlefish (Sepia officinalis L.): the morphology and development of body patterns

[28] and their relation to behaviour," Philosophical Transactions of The Royal Society B: Biological Sciences. vol. 320, pp. 437–487, 1988.

[29] S. E. Temple et al., "High-resolution polarisation vision in a cuttlefish," Current biology: CB, vol. 22, no. 4, R121-2, 2012.

[30] J. M. Ureña Gómez-Moreno, "The 'Mimic' or 'Mimetic' Octopus? A Cognitive-Semiotic Study of Mimicry and Deception in Thaumoctopus Mimicus," Biosemiotics, vol. 12, no. 3, pp. 441–467, 2019.

[31] A. Laan, T. Gutnick, M. J. Kuba, and G. Laurent, "Behavioral analysis of cuttlefish traveling waves and its implications for neural control," Current biology, vol. 24, no. 15, pp. 1737–1742, 2014.

[32] R. L. Caldwell, R. Ross, A. Rodaniche, and C. L. Huffard, "Behavior and Body Patterns of the Larger Pacific Striped Octopus," PloS one, vol. 10, no. 8, 2015.

[33] A. Schnell and G. Vallortigara, "'Mind' is an ill-defined concept: Considerations for future cephalopod research," Animal Sentience, vol. 4, no. 26, 2019.

[34] "Etic Lab, ISCRI: An AI Programmed By An Octopus" [Online]. Available: https://eticlab.co.uk/an-artificial-intelligence-programmed-by-an-octopus-iscri/ (accessed: 12-Mar-21).

[35] "Intelligent robotic octopus settles in at IB". [Online]. Available: https://www.mtab.eu/case/intelligent-robotic-octopus-settles-in-at-ibm/ (accessed: 12-Mar-21).

[36] "Octopus Integrating Project" [Online]. Available: http://www.octopus-project.eu/about.html (accessed: 12-Mar-21).

[37] "Octopus tentacles inspire better prosthetics for humans" Michigan State University, 17-Jun-20. https://msutoday.msu.edu/news/2020/octopus-tentacles-inspire-better-prosthetics-for-humans (accessed: 10-Mar-21).

[38] G. Pickard, "Decoding the mind of an octopus," University of Nevada, Reno, 25-Oct-18, https://www.unr.edu/nevada-today/news/2018/gideon-caplovitz-awarded-nsf-funds (accessed: 17-Mar-21).

[39] T. Li, "Learning from the octopus: sensorimotor control of octopus-inspired soft robots," University of Zurich, 2013.

[40] T. G. Thuruthel, "Machine Learning Approaches for Control of Soft Robots," Unpublished, 2019.

[41] B. Marenko, "FutureCrafting. A Speculative Method for an Imaginative AI." AAAI Spring Symposia, 2018.

[42] "The Octopus Project | Neuroscience Program at Illinois", [Online]. Available: https://neuroscience.illinois.edu/octopus-project (accessed: 12-Mar-21).

[43] A. Íñiguez, "The Octopus as a Model for Artificial Intelligence - A Multi-Agent Robotic Case Study," in Proceedings of the 9th International Conference on Agents and Artificial Intelligence, Porto, Portugal, 2242017, pp. 439–444.

[44] E. Olajubu, A. Akinwale, and K. Ogundoyin, "An octopus-inspired intrusion deterrence model in distributed computing system", vol. 17, no. 4, p. 483, 2016.

[45] M. Helmstaedter, "The mutual inspirations of machine learning and neuroscience," Neuron, vol. 86, no. 1, pp. 25–28, 2015.

[46] R. R. Murphy, Introduction to AI robotics. Cambridge, MIT Press, 2005.

[47] K. Chin, T. Hellebrekers, and C. Majidi, "Machine Learning for Soft Robotic Sensing and Control," Advanced Intelligent Systems, vol. 2, no. 6, p. 1900171, 2020.

[48] S. Kim, C. Laschi, B. Trimmer, "Soft robotics: a bioinspired evolution in robotics," Trends in biotechnology, vol. 31, no. 5, pp. 287–294, 2013.

[49] R. Pfeifer, M. Lungarella, and F. Iida, "Self-organization, embodiment, and biologically inspired robotics," Science (New York), vol. 318, no. 5853, pp. 1088–1093, 2007.

[50] M. Calisti et al., "An octopus-bioinspired solution to movement and manipulation for soft robots," Bioinspiration & biomimetics, vol. 6, no. 3, p. 36002, 2011.

[51] T. Li, K. Nakajima, M. J. Kuba, T. Gutnick, and R. Pfeifer, "From the octopus to soft robot control: An octopus inspired behaviour control architecture for soft robots," Vie et Milieu, vol. 61, pp. 211–217, 2012.

[52] B. Mazzolai, L. Margheri, M. Cianchetti, P. Dario, and C. Laschi, "Soft-robotic arm inspired by the octopus: II. From artificial requirements to innovative technological solutions," Bioinspiration & biomimetics, vol. 7, no. 2, p. 25005, 2012.

[53] C. Laschi, et al., "Soft Robot Arm Inspired by the Octopus," Advanced Robotics, vol. 26, no. 7, pp. 709–727, 2012.

[54] M. Cianchetti, M. Calisti, L. Margheri, M. Kuba, and C. Laschi, "Bioinspired locomotion and grasping in water: the soft eight-arm Octopus robot," Bioinspiration & biomimetics, vol. 10, no. 3, 2015.

[55] Z. Xie et al., "Octopus arm-inspired tapered soft actuators with suckers for improved grasping," Soft Robotics, 2020.

[56] M. Calisti et al., "Study and fabrication of bioinspired Octopus arm mockups tested on a multipurpose platform," 3rd IEEE RAS and EMBS International Conference on Biomedical Robotics and Biomechatronics (BioRob 2010), 26-29 September 2010, Tokyo, Japan, pp. 461–466.

[57] I. Zelman, et al., "Kinematic decomposition and classification of octopus arm movements," Frontiers in computational neuroscience, vol. 7, p. 60, 2013.

[58] S. Kuroe and K. Ito, "Autonomous Control of Octopus-Like Manipulator Using Reinforcement Learning," in Distributed Computing and Artificial Intelligence: 9th International Conference, Berlin, Heidelberg, 2012, pp. 553–556.

[59] B. G. Woolley and K. O. Stanley, "Evolving a Single Scalable Controller for an Octopus Arm with a Variable Number of Segments," in Parallel Problem Solving from Nature ' PPSN XI, Berlin, Heidelberg, 2011, pp. 270–279.

[60] F. Tramacere, M. Follador, N. M. Pugno, and B. Mazzolai, "Octopus-like suction cups: from natural to artificial solutions," Bioinspiration & biomimetics, vol. 10, no. 3, p. 35004, 2015.

[61] S. Chun et al., "Water-Resistant and Skin-Adhesive Wearable Electronics Using Graphene Fabric Sensor with Octopus-Inspired Microsuckers," ACS applied materials & interfaces, vol. 11, no. 18, pp. 16951–16957, 2019.

[62] R. R. Da Fonseca et al., "A draft genome sequence of the elusive giant squid, Architeuthis dux," GigaScience, vol. 9, no. 1, 2020.

[63] M. Wang and E. Mohlhenrich, Evolutionary Analyses of RNA Editing and Amino Acid Recoding in Cephalopods, 2019.

[64] L. M. Mäthger, S. B. Roberts, and R. T. Hanlon, "Evidence for distributed light sensing in the skin of cuttlefish, Sepia officinalis," Biology letters, vol. 6, no. 5, pp. 600–603, 2010.

[65] M. D. Ramirez and T. H. Oakley, "Eye-independent, light-activated chromatophore expansion (LACE) and expression of phototransduction genes in the skin of Octopus bimaculoides," The Journal of experimental biology, vol. 218, Pt 10, pp. 1513–1520, 2015.

[66] A. Chatterjee, et al., "Cephalopod-inspired optical engineering of human cells," Nature communications, vol. 11, no. 1, p. 2708, 2020.

[67] A. Fishman, S. Catsis, M. Homer, and J. Rossiter, "Touch and see: Physical interactions stimulating patterns in artificial cephalopod skin," IEEE International Conference on Soft Robotics (RoboSoft): 24-28 April 2018, Livorno, pp. 1–6.

[68] "Imnovation, This Synthetic Skin Has Octopus-like Camouflage Capabilities," [Online]. Available: https://www.imnovation-hub.com/science-and-technology/synthetic-skin-octopus-camouflage-capabilities/ (accessed: 12-Mar-21).

[69] C. Yu et al., "Adaptive optoelectronic camouflage systems with designs inspired by cephalopod skins," PNAS, vol. 111, no. 36, pp. 12998–13003, 201.

[70] P. Hanaphy, "Rutgers engineers 3D print octopus-inspired camouflage-ready robots," [Online]. https://3dprintingindustry.com/news/rutgers-engineers-3d-print-octopus-inspired-camouflage-ready-robots-181744/ (accessed: 17-Mar-21).